\documentclass[a4paper]{svproc}
\pdfoutput=1

\usepackage{url}
\usepackage{amsmath}
\newtheorem{assumption}{Assumption}
\usepackage{amssymb}
\usepackage{graphicx}

\usepackage{algorithm,algpseudocode}

\usepackage{wrapfig}

\newcommand{\ie}{\textrm{i.e.}\xspace}

\newcommand{\sref}{Section~\ref}

\newcommand{\aref}[1]{Algorithm~\ref{#1}}
\newcommand{\eref}[1]{eq.(\ref{#1})}
\newcommand{\fref}[1]{Figure~\ref{#1}}

\newcounter{comment}

\newcommand{\pomdp}{\ensuremath{P}\xspace}

\newcommand{\transF}{\ensuremath{T}\xspace}
\newcommand{\transFComp}{\ensuremath{T(s, a, s')}\xspace}

\newcommand{\obsF}{\ensuremath{Z}\xspace}

\newcommand{\rewFunc}{\ensuremath{R}\xspace}

\newcommand{\policy}{\ensuremath{\pi}\xspace}

\newcommand{\optPol}{\ensuremath{\pi^*}\xspace}

\newcommand{\stSpace}{\ensuremath{\mathcal{S}}\xspace}

\newcommand{\st}{\ensuremath{s}\xspace}
\newcommand{\stp}{\ensuremath{s'}\xspace}
\newcommand{\actSpace}{\ensuremath{\mathcal{A}}\xspace}
\newcommand{\act}{\ensuremath{a}\xspace}
\newcommand{\obsSpace}{\ensuremath{\mathcal{O}}\xspace}
\newcommand{\obs}{\ensuremath{o}\xspace}

\newcommand{\bel}{\ensuremath{b}\xspace}

\newcommand{\alg}{MLPP\xspace}
\newcommand{\algLong}{Multilevel POMDP Planner\xspace}

\newcommand{\belTree}{\ensuremath{\mathcal{T}}\xspace}

\newcommand{\expect}[1]{\ensuremath{\mathbb{E}[#1]}\xspace}
\newcommand{\app}[1]{\ensuremath{\widehat{#1}}}
\newcommand{\var}[1]{\ensuremath{\mathbb{V}\left [#1 \right ]}}

\newcommand{\hist}{\ensuremath{h}\xspace}
\newcommand{\ep}{\ensuremath{e}\xspace}

\newcommand{\scene}[1]{\texttt{#1}}

\algnewcommand{\IIf}[1]{\State\algorithmicif\ #1\ \algorithmicthen}
\algnewcommand{\EndIIf}{\unskip\ \algorithmicend\ \algorithmicif}
\algnewcommand{\IfThenElse}[3]{
  \State \algorithmicif\ #1\ \algorithmicthen\ #2\ \algorithmicelse\ #3}

\usepackage{xspace}

\DeclareMathOperator*{\argmax}{arg\,max}
\graphicspath{ {img/} }

\begin{document}
\mainmatter              
\title{Multilevel Monte-Carlo for Solving POMDPs Online}
\titlerunning{Multilevel POMDP Planner}  
%
\author{Marcus Hoerger\inst{1} \and Hanna Kurniawati\inst{1} \and
Alberto Elfes\inst{2}}
\authorrunning{Marcus Hoerger et al.} 
%
\tocauthor{Marcus Hoerger, Hanna Kurniawati, Alberto Elfes}
\institute{Research School of Computer Science, Australian National University, ACT, Australia,\\
\email{\{marcus.hoerger, hanna.kurniawati\}@anu.edu.au}
\and
Robotics and Autonomous Systems Group, Data61, CSIRO,\\
\email{Alberto.Elfes@data61.csiro.au}}

\maketitle              

\begin{abstract}
Planning under partial obervability is essential for autonomous robots. A principled way to address such planning problems is the Partially Observable Markov Decision Process (POMDP). Although solving POMDPs is computationally intractable, substantial advancements have been achieved in developing approximate POMDP solvers in the past two decades.
However, computing robust solutions for systems with complex dynamics remain challenging. Most on-line solvers rely on a large number of forward-simulations and standard Monte-Carlo methods to compute the expected outcomes of actions the robot can perform. For systems with complex dynamics, e.g., those with non-linear dynamics that admit no closed form solution, even a single forward simulation can be prohibitively expensive. Of course, this issue exacerbates for problems with long planning horizons. This paper aims to alleviate the above difficulty. To this end, we propose a new on-line POMDP solver, called \algLong (\alg), that combines the commonly known Monte-Carlo-Tree-Search with the concept of Multilevel Monte-Carlo to speed-up our capability in generating approximately optimal solutions for POMDPs with complex dynamics. Experiments on four different problems of POMDP-based torque control, navigation and grasping indicate that \alg substantially outperforms state-of-the-art POMDP solvers.
\end{abstract}

\section{Introduction}\label{sec:intro}
Planning under partial observability is both challenging and essential for autonomous robots. 
To operate reliably, an autonomous robot must act strategically to accomplish its tasks, despite being subject to various types of uncertainties, such as motion and sensing uncertainty, and uncertainty regarding the environment the robot operates in. Due to these uncertainties, the robot does not have full observability on the state of the robot and/or its operating environment. 
The Partially Observable Markov Decision Processes (POMDP)\cite{Sondik:71} is a mathematically principled way to solve such planning problems. 

Although solving a POMDP exactly is computationally intractable\cite{papadimitriou1987complexity}, the past two decades have seen tremendous progress in developing approximately optimal solvers that trade optimality for computational tractability. Various solvers have been proposed for POMDPs with large state spaces\cite{kurniawati2016online,Kurniawati08sarsop:efficient,luo2016importance,Pin03:Point,silver2010monte,Smi05:Point,somani2013despot,sunberg2018online}, large observation spaces\cite{bai2014integrated,Hoey2005}, large or continuous actions spaces\cite{seiler2015online,Wang2018} and long planning horizons\cite{agha2011firm,He2010:Puma,kurniawati2011motion}, enabling POMDPs to start to become practical for various robotics planning problems\cite{bai2012unmanned,horowitz2013interactive,hsiao2007grasping}. 

Most state-of-the-art on-line solvers, such as POMCP\cite{silver2010monte}, DESPOT\cite{somani2013despot}, and ABT\cite{kurniawati2016online} rely on a large number of forward simulations of the system and standard Monte-Carlo to estimate the expected values of different sequences of actions. While this strategy has substantially improved state-of-the-art solvers, 
their performance degrades for problems with complex non-linear dynamics where even a one-step forward simulation requires expensive numerical integrations.
Aside from complex dynamics, long planning horizon problems ---that is, problems that require more than 10 look-ahead steps before a good solution can be found--- remain challenging for on-line solvers. In such problems, even when the computational cost for a one-step forward simulation is cheap, the solver must evaluate long sequences of actions before a good solution is found. 

Although complex dynamics and long planning horizons seem like separate issues, both can be alleviated via simplified dynamics. For instance, simplifying the dynamics to reduce the cost of a one-step forward simulation would alleviate the first issue, while simplifying the dynamics, so as to reduce the amount of control inputs switching, could reduce the effective planning horizon. 
Simplified dynamics models are widely used in deterministic planning and control, albeit less so in solving POMDPs. 

In this paper we propose a sampling-based on-line POMDP solver, called \algLong (\alg), that uses multiple levels of approximation to the system's dynamics to reduce the number and complexity of forward simulations needed to compute a near-optimal policy. 
\alg combines the commonly used Monte-Carlo-Tree-Search\cite{kocsis2006bandit} with a relatively recent concept in Monte-Carlo, called Multilevel Monte-Carlo (MLMC)\cite{giles2015multilevel,hein01:multilevel}. MLMC is a variance reduction technique that uses cheap and coarse approximations of the system to carry out the majority of the simulations and combines them with a small number of accurate but expensive simulations to maintain correctness. 
By constructing a set of correlated  samples from a sequence of approximations of the original system's dynamics, in conjunction with applying Multilevel Monte-Carlo estimation to compute the expected value of sequences of actions, 
\alg is able to compute near-optimal policies substantially faster than two of the fastest today's on-line solvers on four challenging robotic planning tasks under uncertainty. Two of these scenarios are articulated robots with POMDP-based torque control, while the other two have a required planning horizon of more than 10 steps. We also show that under certain conditions, \alg converges asymptotically to the optimal solution.  

\section{Background}\label{sec:background}
\subsection{Partially Observable Markov Decision Process (POMDP)}\label{ssec:POMDP}
Formally a POMDP is a tuple $<\stSpace, \actSpace, \obsSpace, \transF, \obsF, \rewFunc, \gamma>$, where \stSpace, \actSpace and \obsSpace are the state, action and observation spaces of the robot. \transF and \obsF model the uncertainty in the effect of taking actions and perceiving observations as conditional probability functions $\transFComp = p(\stp | \st, \act)$ and $Z(\stp, \act, \obs) = p(\obs | \stp, \act)$, where $\st, \stp \in \stSpace$, $\act \in \actSpace$ and $\obs \in \obsSpace$. $R(\st, \act)$ models the reward the robot receives when performing action $\act$ from $\st$ and $0 < \gamma < 1$ is a discount factor. Due to uncertainties in the effect of executing actions and perceiving observations, the true state of the robot is only partially observable. Thus, given a \textit{history} $\hist_t = \{\act_0, \obs_0, ..., \act_t, \obs_t\}$ of previous actions and observations, the robot maintains a \textit{belief} $\bel(\st, \hist_t)$, a probability distribution over states, conditioned on history $\hist_t$, and selects actions according to a \textit{policy} $\pi(\hist_t)$, a mapping from histories to actions. The value of a policy $\pi$ is the expected discounted future reward the robot receives when following $\pi$ given \hist, i.e. $V_{\pi}(\hist) = \sum_{t=0}^{\infty}\gamma^t \expect{r_t | \hist, \pi}$. The solution of a POMDP is then an optimal policy \optPol such that $\optPol = \argmax_{\pi} V_{\pi}(\hist)$.

\subsection{Multilevel Monte-Carlo}
Since its introduction in 2001, MLMC has been used to significantly reduce the computational effort on applications that involve computing expectations from expensive simulations\cite{giles2008multilevel,bierig2016approximation,anderson2012multilevel}. Here, we provide a brief overview of the underlying concept of MLMC. An extensive overview is available at\cite{giles2015multilevel}.  

Suppose we have a random variable $X$ and we wish to compute its expectation $\expect{X}$. A simple Monte-Carlo (MC) estimator for $\expect{X}$ is $\expect{X} \approx \frac{1}{N}\sum_{i=1}^{N}X^{(i)}$, where $X^{(i)}$ are iid. samples drawn from $X$. In many applications sampling from $X$ directly is expensive, causing the MC-estimator to converge slowly.

The idea of MLMC is to use the linearity of expectation property to reduce the cost of sampling. Suppose, $X_0, X_1, X_2, \ldots, X_L$ is a sequence of approximations to $X$, where $\lim_{L\to\infty} X_L = X$ and the approximation increases in accuracy and sampling cost as the index increases. Using the linearity of expectation, we have the simple identity:
\begin{equation}
    \expect{X} = \expect{X_0} + \sum_{l=1}^{L}\expect{X_{l} - X_{l-1}}
\end{equation}
and can design the unbiased estimator:
\begin{equation}\label{eq:mlmcEstimator}
\expect{X} \approx \frac{1}{N_0} \sum_{i = 1}^{N_0}X_0^{(i)} + \sum_{l=1}^{L}\frac{1}{N_l}\sum_{i=1}^{N_l}(X_{l}^{(i)} - X_{l-1}^{(i)})
\end{equation}
with independent samples at each level. The key here is that even though the samples at each level are independent, the individual samples $X_l^{(i)}$ and $X_{l-1}^{(i)}$ at level $l$ are correlated, such that their differences have a small variance. Of course, the aim is to be able to sample only from the first few approximations while still computing a relatively good approximation of $\expect{X}$. It turns out, if we define the sequence of approximations appropriately\cite{giles2015multilevel}, the variance $\var{X_{l} - X_{l-1}}$ becomes smaller for increasing level $l$, and therefore we require fewer and fewer samples to accurately estimate the expected differences. This means we can take the majority of the samples at the coarser levels, where sampling is cheap, and only a few samples are required on the finer levels, thereby leading to a substantial reduction of the cost to estimate $\expect{X}$ accurately.

\section{\algLong (\alg)}\label{sec:mlmcPomdp}
\alg is an anytime on-line POMDP solver. Starting from the current history $\hist_t$, \alg computes an approximation to the optimal policy by iteratively constructing and evaluating a search tree \belTree, a tree whose nodes are histories and edges represent a pair of action-observation. From hereafter, we use the term \textit{nodes} and the \textit{histories} they represent interchangeably. 
A history $\hist'$ is a child node of $\hist$ via edge $(\act, \obs)$ if $\hist' = \hist\act\obs$. The root of \belTree corresponds to an empty history $\hist_0$. 
The policy of \alg is embedded in \belTree via 
$\policy(\hist) = \argmax_{\act \in \actSpace} \widehat{Q}(\hist, \act)$, 
where $\widehat{Q}(\hist, \act)$ is an approximation of $Q(\hist, \act) = R(\hist, \act) + \gamma \mathbb{E}_{\obs \in \obsSpace}[V_{\pi^*}(\hist\act\obs)]$, \ie the expected value of executing \act from \hist and continuing optimally afterwards.

To compute $\widehat{Q}$, \alg constructs \belTree using a framework similar to POMCP\cite{silver2010monte} and ABT\cite{kurniawati2016online}: Given the current history $\hist_t$, \alg repeatedly samples \textit{episodes} starting from $\hist_t$. An episode \ep is a sequence of $(\st, \act, \obs, r)$-quadruples, where the state $\st \in \stSpace$ of the first quadruple is distributed according to the current belief $\bel_t$ -- we approximate beliefs by sets of particles -- and the states of all subsequent quadruples are sampled from the transition function $T$, given the state and action of the previous quadruple. The observations $\obs \in \obsSpace$ are sampled from the observation function $Z$, while the reward $r = R(\st, \act)$ is generated by the simulation process. Each episode corresponds to a path in \belTree. Details on how the episodes are sampled are given in \sref{ssec:samplingHistories}. 

Key to \alg is the adoption of the MLMC concept: Episodes are sampled using multiple levels of approximations of the transition function. Suppose $T$ is the transition function of the POMDP problem. \alg first defines a sequence of increasingly accurate approximations of the transition function $T_0, T_1, ..., T_L$ with $T_L = T$, and uses the less accurate but cheaper transition functions for the majority of the episode samples, to approximate the $Q$-value function fast. 
Note that to ensure asymptotic convergence of \alg, we slightly modify MLMC such that $L$ is finite and $T_L$ is the most refined level \alg samples from. 

Let 
$V_k(\ep)$ be the total discounted reward of an episode starting from the $k$-th quadruple. For a node \hist of depth $k$, \alg approximates $Q(\hist, \act)$ according to:
\begin{align}\label{eq:mlmcQEstimator}
\widehat{Q}(\hist, \act) &= \widehat{Q}_0(\hist, \act) + \sum_{l=1}^{L} (\widehat{Q}_{l}(\hist, \act) - \widehat{Q}_{l-1}(\hist, \act)) \nonumber \\
= \frac{1}{N_0(\hist, \act)} & \sum_{i=1}^{N_0(\hist, \act)} V_k(\ep_0^{(i)}) + \sum_{l = 1}^{L} \frac{1}{N_l(\hist, \act)} \sum_{i=1}^{N_l(\hist, \act)}(V_k(\ep_{l}^{(i)}) - V_k(\ep_{l-1}^{(i)}))
\end{align}
where an episode $\ep_l$ on level $l$ is sampled using $T_l$, and $N_l(\hist, \act)$ is the number of all episodes on level $l$ that start from $\hist_0$, pass through $\hist$ and execute \act from $\hist$. Similar to  \eref{eq:mlmcEstimator}, the key here is that even though we draw independent samples on each level, the episode samples for the value differences $V_k(\ep_{l}^{(i)}) - V_k(\ep_{l-1}^{(i)})$ are \textit{correlated}. The question is, \emph{how do we correlate the sampled episodes?}

We adopt the concepts of \textit{determinization}\cite{somani2013despot} and common random numbers \cite{mcbook}, a popular variance reduction technique: To sample states and observations for an episodes on level $l$, we use a deterministic simulative model, i.e. a function $f_{l}: \stSpace \times \actSpace \times \left [0, 1 \right ] \mapsto \stSpace \times \obsSpace$ such that, given a random variable $\psi$ uniformly distributed in $\left [0, 1 \right ]$, $(\stp, \obs) = f_{l}(\st, \act, \psi)$ is distributed according to $T_{l}(\st, \act, \stp)O(\stp, \act, \obs)$. For an initial state $\st_0 \sim \bel_t$ and a sequence of actions, the states and observations of an episode on level $l$ are then \textit{deterministically} generated from $f_l$ using a sequence $\Psi = (\psi_0, \psi_1, ...)$ of iid. random numbers. Now, to sample a correlated episode on level $l-1$, we use the same initial state $s_0$, the same sequence of actions and the same random sample $\Psi$ used for the episode on level $l$, but generate next states and observations from the model $f_{l-1}$ corresponding to $T_{l-1}$, such that for a given $\st$ and $\act$, $(\stp, \obs) = f_{l-1}(\st, \act, \psi)$ is distributed according to $T_{l-1}(\st, \act, \stp)O(\stp, \act, \obs)$. Using the same initial state, action sequence and random sample $\Psi$ results in two closely correlated episodes, reducing the variance of $V_k(\ep_{l}^{(i)}) - V_k(\ep_{l-1}^{(i)})$.

To incorporate the above sampling strategy to the construction of \belTree, \alg computes the estimator \eref{eq:mlmcQEstimator} in two subsequent stages: In the first stage, \alg samples episodes using the coarsest approximation $\transF_0$ of the transition function to compute the first term in \eref{eq:mlmcQEstimator}. In the second stage, \alg samples correlated pairs of episodes to compute the value difference terms in \eref{eq:mlmcQEstimator}. These two stages are detailed in the next two subsections. An overview of \alg is shown in \aref{alg:mlmc-abt}, procedure \textproc{RunMLPP}. We start by initializing \belTree, containing the empty history $\hist_0$ as the root, and setting the current belief to be the initial belief (line 1). Then, in each planning loop iteration (line 3-7) we first sample an episode using $T_0$ (line 4), followed by sampling two correlated episodes (line 6). Once the planning time for the current step is over, \alg executes the action that satisfies $\argmax_{\act \in \actSpace} \widehat{Q}(\hist_0, \act)$. Based on the executed action \act and perceived observation \obs, we update the belief using a SIR particle filter\cite{arulampalam2002tutorial} (line 11) and continue planning from the updated history $\hist_0\act\obs$. This process repeats until a maximum number of steps is reached, or the system enters a terminal state (we assume that we know when the system enters a terminal state).
\begin{algorithm}[H]
\begin{tabular}{c}
\begin{minipage}{\textwidth}
\textproc{RunMLPP}
\begin{algorithmic}[1]
\State $\mathcal{T} = $\ initializeTree(); $\bel = \bel_0$; $\hist = $\ Root of \belTree; terminal\ =\ False; $t=1$
\While{terminal is False and $t < t_{max}$}
   \While{planning time not over}
       \State $(\ep, \Psi)$ = SampleEpisode($\mathcal{T}$, \bel, $\hist$, 0)
       \State backupEpisode(\belTree, \ep)
       \State SampleCorrelatedEpisodes($\mathcal{T}$, \bel, $\hist$)
   \EndWhile
   \State $\act =$ get best action in $\mathcal{T}$ from $\hist$
   \State terminal = Execute $\act$
   \State $o = $\ get observation
   \State $\bel = \tau(\bel, \act, \obs)$; $\hist = \hist\act\obs$
   \State $t=t+1$
\EndWhile
\end{algorithmic}
\end{minipage} \\ [0.1ex]  \\
\begin{minipage}{\textwidth}
\textproc{SampleEpisode}(Search tree $\mathcal{T}$, Belief $\bel$, History node \hist, level $l$)
\begin{algorithmic}[1]
\State $\st = $\ sample a state from \bel
\State $\ep = $\ init episode; $\Psi = $ init random number sequence; unvisitedAction = False
\While{unvisitedAction is False and \st not terminal}
  \State $(\act,$\ unvisitedAction$) = $\ UCB1$(\hist, l)$ \Comment{For $l = 0$ we select actions from \actSpace, for $l > 0$ from $\actSpace'(\hist)$}
  \IIf{\act is $\varnothing$} break \EndIIf
  \State $\psi \sim [0, 1]$
  \State $(\stp, \obs) = f_l(\st, \act, \psi)$ \Comment{Generate $(\stp, \obs)$ such that $(\stp, \obs) \sim T_l(\st, \act, \stp)Z(\stp, \act, \obs)$}
  \State $r = R(\st, \act)$; insert $(\st, \act, \obs, r)$ to $\ep$ and $\psi$ to $\Psi$  
  \State $\st = \stp$; $\hist = $ child node of \hist via edge $(\act, \obs)$. If no such child exists, create one  
\EndWhile
\State $r = 0$
\IIf{unvisitedAction is True} $r = $\ calculateHeuristic(\st, \hist) \EndIIf
\State insert $(s, -, -, r)$ to $\ep$
\State \Return $(\ep, \Psi)$
\end{algorithmic}
\end{minipage} \\ [0.1ex]  \\
\begin{minipage}{\textwidth}
\textproc{SampleCorrelatedEpisodes}(Search tree \belTree, Belief \bel, History node \hist)
\begin{algorithmic}[1]
\State $l \sim 2^{-l}$\Comment{Sample a level $l$ proportional to $2^{-l}$}
\State $(\ep_{l}, \Psi) = $\ sampleEpisode($\belTree, \bel, \hist, l$)
\State $\ep_{l-1} =$\ init episode
\State $\st = \ep_l[1].\st$ \Comment{State of the first quadruple of $\ep_l$} 
\For{$i = 1$ to $\left |h_l \right |$}
\State $\act = \ep_{l}[i].\act$ \Comment{Action of the $i$-th quadruple of $\ep_l$}
\State $(\stp, \obs) = f_{l-1}(\st, \act, \Psi[i])$ \Comment{\stp is generated according to $\transF_{l-1}$}
\State $r = R(\st, \act)$; insert $(\st, \act, \obs, r)$ to $\ep_{l-1}$
\State $\st = \stp$; $\hist = $ child node of \hist via edge $(\act, \obs)$. If no such child exists, create one
\IIf{\stp is terminal} break \EndIIf
\EndFor
\State $r = 0$
\If{$i$\ is\ $\left |\ep_l \right |$}
\State $r = $calculateHeuristic(\st, \hist)
\EndIf
\State insert $(\st, -, -, r)$ to $\ep_{l-1}$ and backupRewardDifference(\belTree, $\ep_l$, $\ep_{l-1}$)
\end{algorithmic}
\end{minipage}
\end{tabular}
\caption{MLPP}\label{alg:mlmc-abt}
\end{algorithm}

\subsection{Sampling the episodes using $T_0$}\label{ssec:samplingHistories}
To sample an episode using $T_0$, starting from the current history \hist, we first sample a state from the current belief which will then correspond to the state of the first quadruple of the episode (line 1 in \aref{alg:mlmc-abt}, procedure \textproc{SampleEpisode}). To sample a next state and observation, we first need to select an action from \hist (line 4). The action-selection strategy is similar to the strategy used in POMCP and ABT. Consider the set of actions $\actSpace'(\hist) \subseteq \actSpace$ that have already been selected from \hist. If $\actSpace'(\hist) = \actSpace$, \ie all actions have been selected from \hist at least once, we formulate the problem of which action to select as a Multi-Arm-Bandit problem (MAB)\cite{Sut12:Reinforcement}. MABs are a class of reinforcement learning problems where an agent has to select a sequence of actions to maximise the total reward, but the rewards of selecting actions is not known in advance. One of the most successful algorithms to solve MAB problems is Upper Confidence Bounds1 (UCB1)\cite{Aue02:Finite}. 
UCB1 selects an action according to $\act = \argmax_{\act \in \actSpace} \left ( \widehat{Q}(\hist, \act) + c_0 \sqrt{\frac{\log(N_0(\hist))}{N_0(\hist, \act)}} \right )$, where $N_0(\hist)$ is the number of episodes that were sampled using $T_0$ that pass through \hist, $N_0(\hist, \act)$ is the number of episodes that were sampled using $T_0$, pass through \hist and select action \act from \hist and $c_0$ is an exploration constant. In case there are actions that haven't been selected from \hist, we use a rollout strategy that selects one of these actions uniformly at random. 

We then sample a random number $\psi \sim [0, 1]$ (line 6) and, based on $\psi$ and the selected action, generate a next state and observation (line 7) from the model $f_0$ using $\transF_0$, an immediate reward (line 8) and add the quadruple to the episode. Additionally we set \hist to the child node that is connected to \hist via the selected action and sampled observation. If this child node doesn't exist yet, we add it to \belTree (line 9). Note that selecting a previously unselected action always results in a new node. To get a good estimate of $\widehat{Q}_0(\hist, \act)$ for a newly selected action, \alg computes a problem dependent heuristic estimate (line 12) in its rollout strategy using the last state of the episode. Computing a heuristic estimate of $\widehat{Q}_0(\hist, \act)$ helps \alg to quickly focus its search on more promising parts of \belTree. 

Once we have sampled the episodes, we backup the expected discounted reward of the episode all the way back to the current history (line 5 in procedure \textproc{RunMLPP}) to update the $\widehat{Q}_0$-values along the selected action sequence.
\subsection{Sampling the correlated episodes}
Once \alg has sampled an episode using the coarsest approximation of $T$, it samples two correlated episodes, via procedure \textproc{SampleCorrelatedEpisodes} in \aref{alg:mlmc-abt}. For this we first sample a level $l$ proportional to $2^{-l}$ (line 1), with $l \geq 1$. This is motivated by the idea that as we increase the level, fewer and fewer samples are needed to get a good estimate of the expected value difference. The idea of randomizing the level is motivated by\cite{rhee2012new}. Based on the sampled level $l$, we first sample an episode using the finer transition function $\transF_{l}$ (line 2). Sampling this episode is similar to the coarsest level, with some notable differences in the action-selection strategy: At each node \hist, we only consider actions from the set $\actSpace'(\hist) \subseteq \actSpace$ that have been selected at least once during sampling of the coarsest episodes. This is because actions that haven't been selected on the coarsest level yet, don't have an estimate for the first component $\widehat{Q}_0(\hist, \act)$ of \eref{eq:mlmcQEstimator}, therefore we wouldn't be able to update the $Q$-value estimates in a meaningful way. Additionally, for each level, we maintain separate visitation counts $N_l(\hist)$ and $N_l(\hist, \act)$, which allows us to use UCB1 as the action selection strategy, i.e. $\act = \argmax_{\act \in \actSpace'(\hist)} \left ( \widehat{Q}(\hist, \act) + c_l \sqrt{\frac{\log(N_l(\hist))}{N_l(\hist, \act)}} \right )$. In case we end up in a node where $\actSpace'(\hist)$ is empty, we stop the sampling of the episode.


To sample a correlated episode on the coarser level $l-1$, we use the model $f_{l-1}$ corresponding to $T_{l-1}$, but the same initial state (line 4), the same action sequence (line 6) and the same random number sequence (line 7) that was used for the episode on level $l$. After we have obtained two correlated episodes on level $l$ and $l-1$, we backpropagate the discounted reward difference between the two episodes along the action sequence all the way to the current history (line 16), to update the expected $Q$-value difference between level $l$ and $l-1$, i.e. $\widehat{Q}_l(\hist, \act) - \widehat{Q}_{l-1}(\hist, \act)$ for each action in the sequence. Note that even though we use the same action sequence for both episodes, the sequence of visited nodes in \belTree might be different due to different observations, or because the coarse episode terminates earlier than fine episode. If this is the case, we backup both episodes individually until we arrive at an action edge that is the same for both episodes (there is always at least one common action edge, which is the outgoing action of the current history). The actual $Q$-value estimates $\app{Q}(\hist, \act)$ along the common action sequence are then updated according to 
\begin{equation}\label{eq:correction_term}
    \app{Q}(\hist, \act) = \app{Q}_0(\hist, \act) + \sum_{l=1}^{K} w_l(\hist, \act) \left ( \widehat{Q}_l(\hist, \act) - \widehat{Q}_{l-1}(\hist, \act) \right )
\end{equation}
During the early stages of planning, when only a few discounted reward differences have been sampled, the estimator of $\widehat{Q}_l(\hist, \act) - \widehat{Q}_{l-1}(\hist, \act)$ might have a large variance, causing it to "overcorrect" the policy. To alleviate this issue, we use a weighting function $w_l$ defined as $w_l(\hist, \act) = \left (1 + \frac{\var{\widehat{Q}_l(\hist, \act) - \widehat{Q}_{l-1}(\hist, \act)}}{N_l} \right )^{-1}$, 
where $\var{\widehat{Q}_l(\hist, \act) - \widehat{Q}_{l-1}(\hist, \act)}$ is an estimate of the variance of $\widehat{Q}_l(\hist, \act) - \widehat{Q}_{l-1}(\hist, \act)$ and $N_l$ is the number of samples used to estimate $\widehat{Q}_l(\hist, \act) - \widehat{Q}_{l-1}(\hist, \act)$. As the number of samples on level $l$ and $l-1$ increases, $w_l(\hist, \act)$ converges towards 1, hence the limit of \eref{eq:correction_term} is the actual MLMC-estimator of $\app{Q}(\hist, \act)$ defined in \eref{eq:mlmcQEstimator}.

\section{Convergence of \alg}\label{sec:discussion}
We now discuss under which conditions \alg converges to the optimal policy. 

Suppose we have an action sequence $(\act_1, \act_2, \act_3, ..., \act_K)$ and an initial state $\st_0 \sim \bel_t$. Applying the action sequence to $\st_0$ results in a \textit{trajectory} $(\st_0,\allowbreak \act_1,\allowbreak \st_1,\allowbreak \obs_1,\allowbreak \act_2,\allowbreak \st_2,\allowbreak \obs_2,\allowbreak ...)$ which is distributed according to $\prod_{i=1}^{K}T(\st_{i-1}, \act_i, \st_i)Z(\st_i, \act_i, \obs_i)$. Now suppose we have a sequence of approximations of the transition function $T_0, T_1, ...T_L$ with $T_L = T$.
\begin{assumption}
Given a POMDP \pomdp, with transition function $T$ and a sequence of approximations of the transition function $T_0, T_1, ...T_L$ with $T_L = T$, then for any action sequence $(\act_1, \act_2, \act_3, ..., \act_K)$, $\prod_{i=1}^{K}T(\st_{i-1}, \act_i, \st_i)Z(\st_i, \act_i, \obs_i) > 0$ implies $\prod_{i=1}^{K}T_l(\st_{i-1}, \act_i, \st_i)Z(\st_i, \act_i, \obs_i) > 0$ for $0 \leq l \leq L$.
\end{assumption}

Intuitively, under this assumption, any node in \belTree than can be reached by episodes that are sampled using the original transition function $T$ can also be reached by episodes that are sampled using $T_l$. Given this assumption, and the fact that we select actions according to UCB1 on each level independently, the estimator $\widehat{Q}(\hist, \act)$ in \ref{eq:correction_term} converges to $Q(\hist, \act)$ in probability as the number of episodes that pass through \hist and execute \act from \hist increases on each level. This is based on the analysis in\cite{silver2010monte,kocsis2006bandit}. Therefore \alg's policy converges to the optimal policy in probability, too. Assumption 1 is quite strong and might be too restrictive for some problems. Relaxing this assumption is subject to future work. Nevertheless, problems whose transition and observation functions for all stat--action pairs are represented as distributions with infinite support (e.g., Gaussian) satisfy the assumption above.


\section{Experiments and Results}\label{sec:experiments}
\begin{figure}
\vspace{-0.5cm}
\centering
\small
\begin{tabular}{c@{\hskip0pt}c@{\hskip0pt}c@{\hskip0pt}c}
\includegraphics[width=0.25\textwidth]{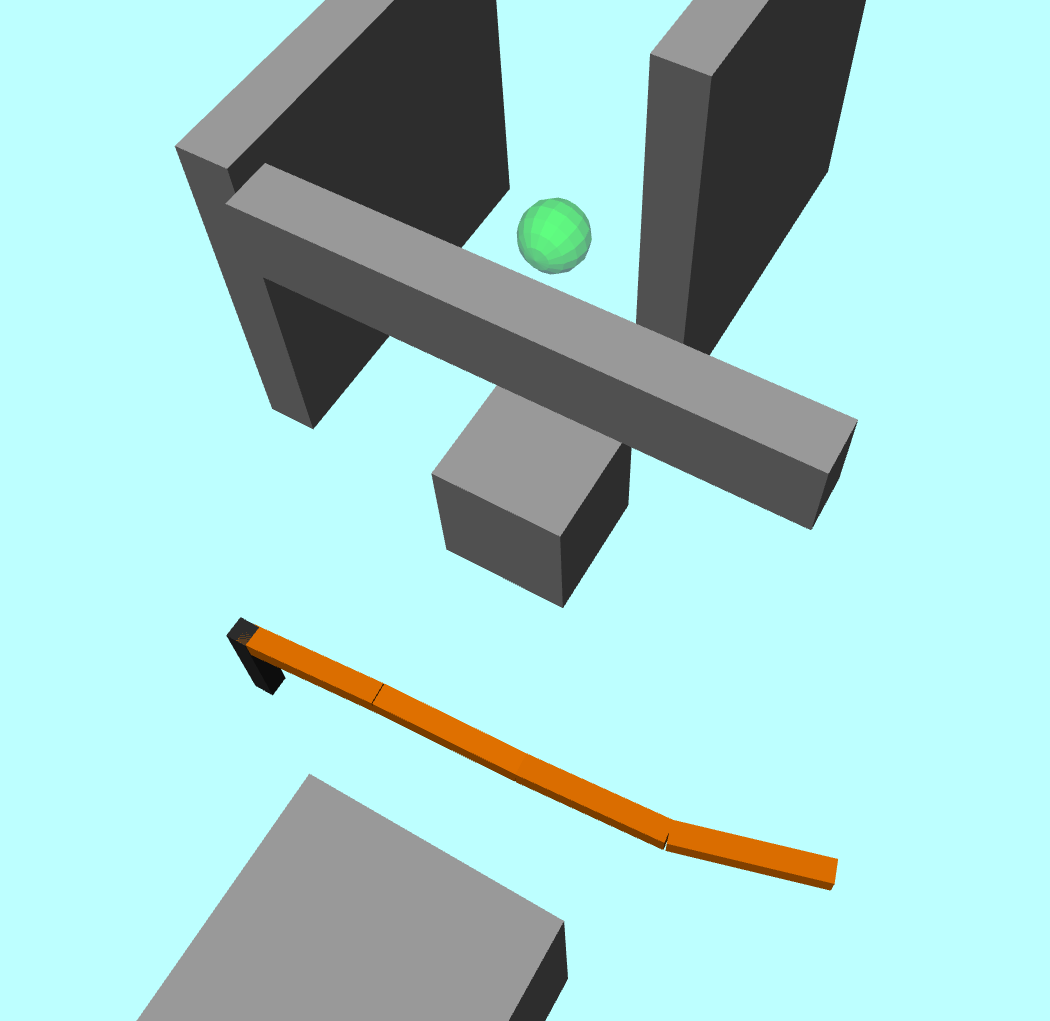} &
\includegraphics[width=0.25\textwidth]{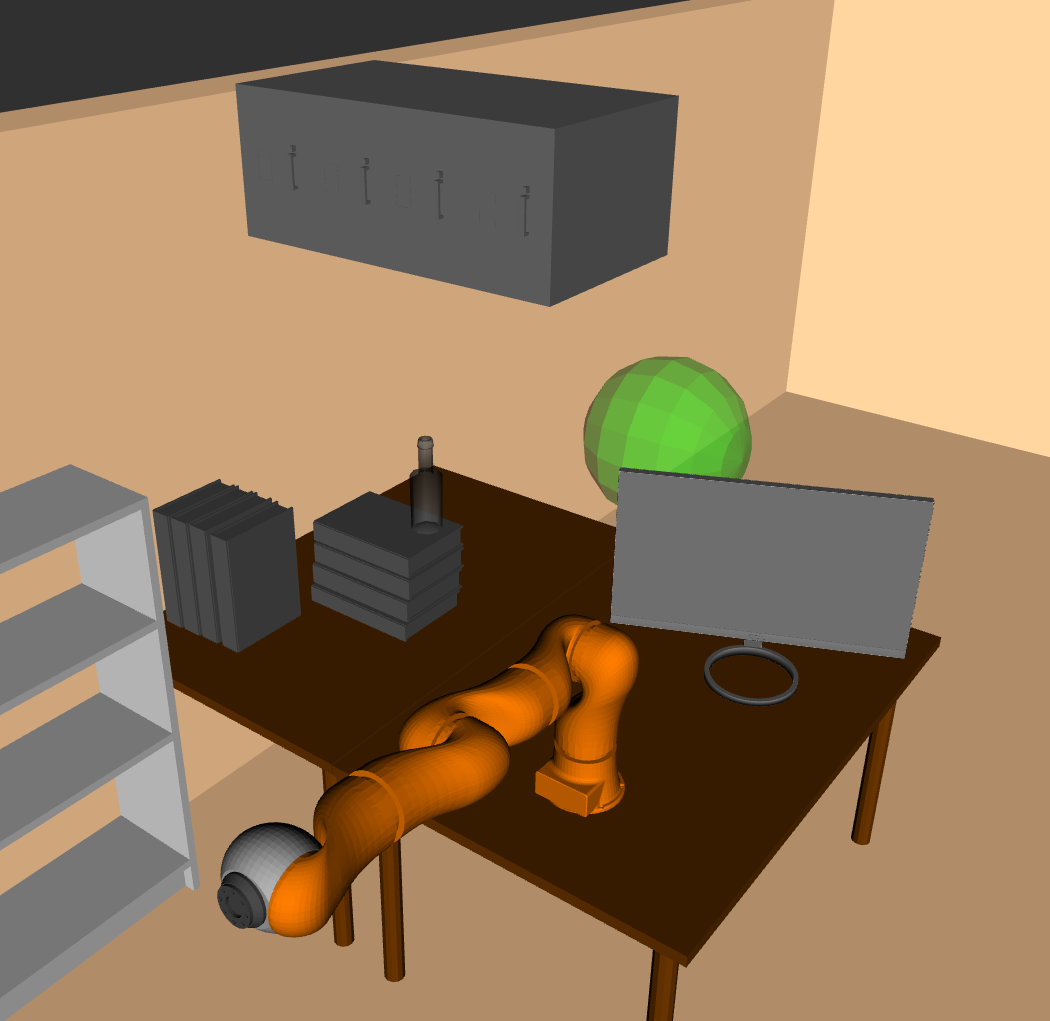} &
\includegraphics[width=0.25\textwidth]{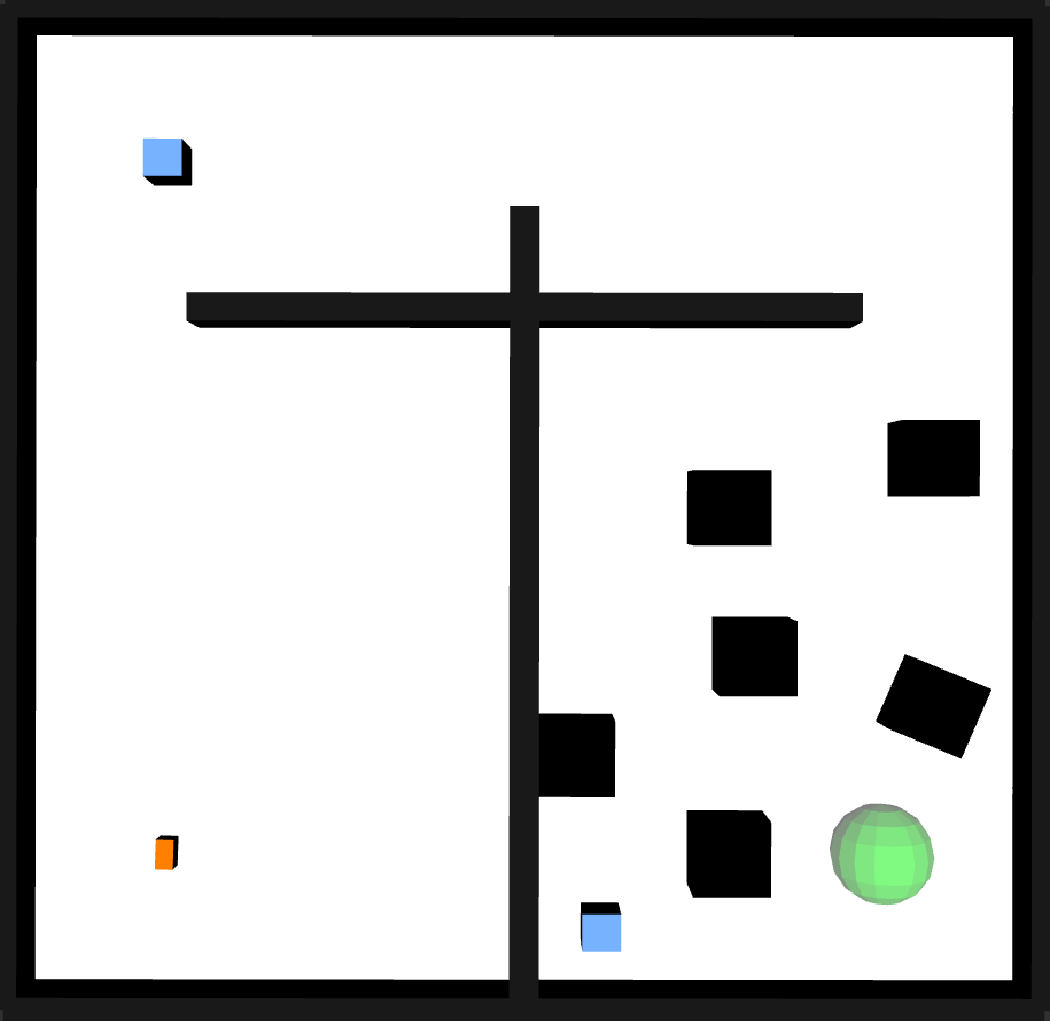} &
\includegraphics[width=0.25\textwidth]{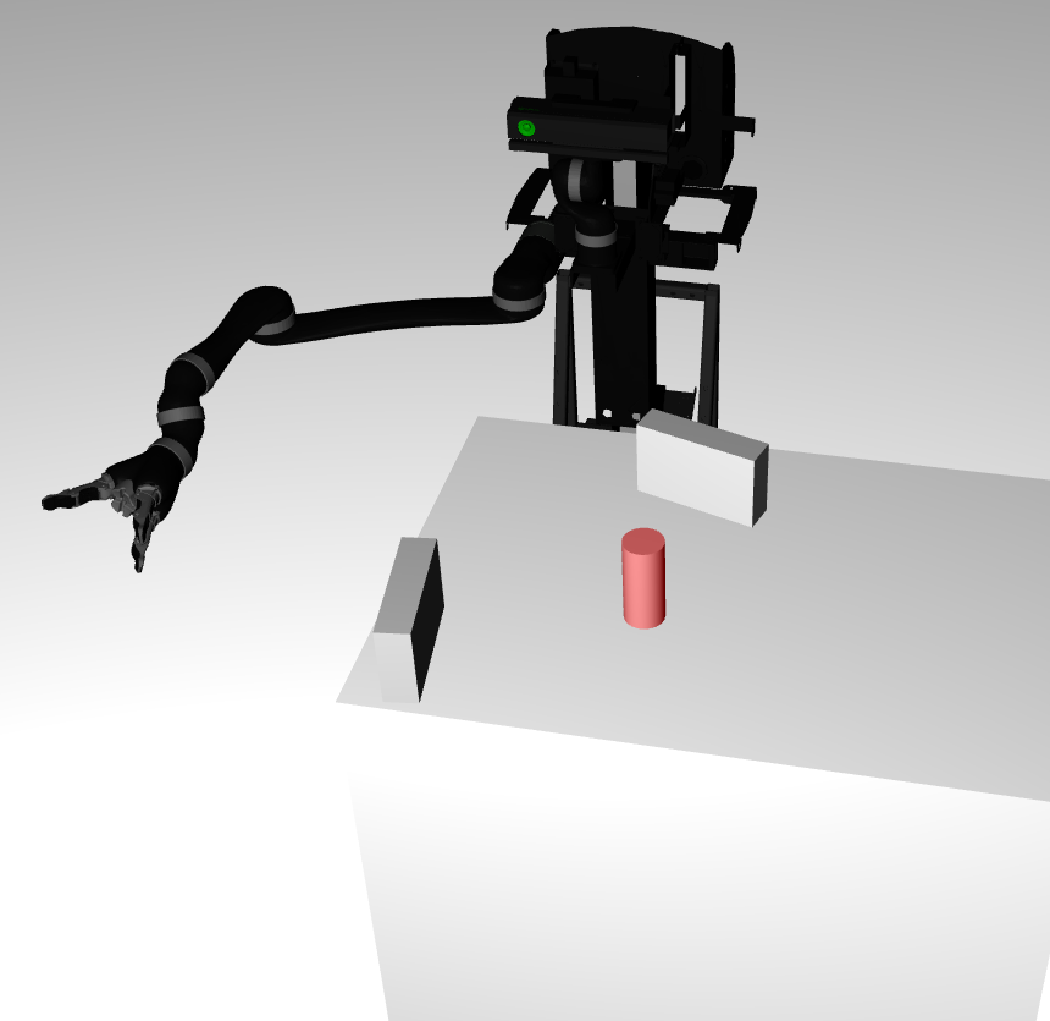} \\
(a) & (b) & (c) & (d)
\end{tabular}
\caption{Test scenarios used to evaluate \alg. (a) \scene{4DOF-Factory} (b) \scene{KukaOffice} (c) \scene{CarNavigation} (d) \scene{MovoGrasping}}
\label{f:scenarios}
\vspace{-0.75cm}
\end{figure}

\vspace{-0.25cm}
\alg is tested on two motion-planning problems under uncertainty with expensive non-linear transition dynamics and two problems with long-planning horizon. 
The scenarios are shown in \fref{f:scenarios} and described below.
\vspace{-0.25cm}
\subsection{Problem scenarios with expensive transition dynamics}\label{ssec:problemScenariosDynamics}
\subsubsection{4DOF-Factory}
A torque-controlled manipulator with 4 revolute joints must move from an initial state to a state where the end-effector lies inside a goal region (colored green in \fref{f:scenarios}(a)), without colliding with any of the obstacles. The state space is the joint product of joint-angles and joint-velocities. The control inputs are the joint-torques. To keep the action space small, the action space is set to be the maximum and minimum possible joint torque, resulting in 16 discrete actions. We assume the input torque at each joint is affected by zero-mean additive Gaussian noise. The dynamics of the manipulators are defined using the well-known Newton-Euler formalism\cite{spong06:RobotModelling}. We assume that each torque input is applied for $\Delta t = 0.1$s. The robot has two sensors: One measures the position of the end-effector inside the robot's workspace, while the other measures the joint velocities. Both measurements are disturbed by zero-mean additive Gaussian noise. The initial state is known exactly, which is when the joint angles and velocities are zero. 

The robot enters a terminal state and receives a reward of 1,000 upon reaching the goal. To encourage the robot to reach the goal area quickly, it receives a small penalty of -1 for every other action. Collision causes the robot to enter a terminal state and receive a penalty of -500. The discount factor is 0.98 and the maximum number of planning steps is limited to 50.
\vspace{-0.5cm}

\subsubsection{KukaOffice}
The scenario is very similar to the \scene{4DOF-Factory} scenario. However, the robot and environment (illustrated in \fref{f:scenarios}(b)) are different. The robot is a Kuka iiwa with 7 revolute joints.
We set the POMDP model to be similar to that of the \scene{4DOF-Factory} scenario, but of course expanded to handle 7DOFs. For instance, the action space remains the maximum and minimum possible joint torque for each joint. However, due to the increase in DOFs, the POMDP model of this scenario has 128 discrete actions. The sensors and errors in both actions and sensing are the same as the \scene{4DOF-Factory} scenario. Similar to the above scenario, we assume each torque input is applied for $\Delta t = 0.1$s. The initial state in this scenario is also similar to the above scenario: The initial joint-velocities are all zero and almost all joint-angles are zero too, except for the second joint, where it is $-1.5rad$.
\vspace{-0.5cm}

\subsection{Problem scenarios with long planning-horizons}\label{ssec:problemScenariosDynamics}
\subsubsection{CarNavigation}
A nonholonomic car-like robot drives on a flat xy-plane inside a 3D environment (shown in Figure \ref{f:scenarios}(c)), populated by obstacles. The robot must drive from a known start state to a position inside the goal region (marked as a green sphere) without colliding with the obstacles. The state of the robot is a 4D-vector consisting of the position of the robot on the $xy$-plane, its orientation $\theta$ around the $z$-axis, and the forward velocity $\upsilon$. The control input is a $2D$-vector consisting of the acceleration $\alpha$ and the steering-wheel angle $\phi_t$. The robot’s dynamics is subject to control noise $v_t = (\tilde{\alpha}_t, \tilde{\phi}_t) \sim N(0, \Sigma_v)$. The transition model of the robot is defined as $\st_{t+1} =\allowbreak \left [x_t + \Delta t \upsilon_t \cos \theta_t;\right .$ $\left . y_t + \Delta t \upsilon_t \sin \theta_t;\right .$ $\left . \theta_t + \Delta t \tan(\phi_t + \tilde{\phi}_t) / 0.11;\right . $ $\left . \upsilon_t + \Delta t (\alpha_t + \tilde{\alpha}_t) \right ]^T$, 
where $\Delta t = 0.05s$ is a fixed parameter that represents the duration of a time-step and the value 0.11 is the distance between the front and rear axles of the wheels. The robot is equipped with two sensors: The first one is a localization sensor that receives a signal from one of two beacons located in the environment (blue squares in Figure \ref{f:scenarios}(c)), with probability proportional to the inverse euclidean distance to the beacons. The second sensor is a velocity sensor mounted on the car. With these two sensors the observation model is $o_t = \left [((x_t - \hat{x})^2 + (y_t - \hat{y})^2 + 1)^{-1}, v_t\right ]^T + w_t$,
where $(\hat{x}, \hat{y})$ is the location of the beacon the robot receives a signal from, $v_t$ is the velocity and $w_t$ is an error vector drawn from a zero-mean multivariate Gaussian distribution. The robot starts from a state where it is located in the lower-left corner of the map. The robot receives a penalty of -500 when it collides with an obstacle, a reward of 10,000 when reaching the goal area (in both cases it enters a terminal state) and a small penalty of -1 for every step. The discount factor is 0.99 and we allow a maximum of 500 planning steps.

For this problem sampling from the transition function is cheap, thanks to the closed-form transition dynamics. However, the robot must perform a large number of steps (around 200) to reach the goal area from its initial state.
\vspace{-0.5cm}
\subsubsection{MovoGrasp}
A 6-DOF Movo manipulator equipped with a gripper must grasp a cylindrical object placed on a table in front of the robot while avoiding collisions with the table and the static obstacles on the table. The environment is shown in Figure \ref{f:scenarios}(d). The state space of the manipulator is defined as $\stSpace = \Theta \times GripperStates \times GraspStates \times \Phi_{obj}$, where $\Theta  = (-3.14rad, 3.14rad)^6$ are the joint angles of the arm, $GripperStates = \{gripperOpen, gripperClosed\}$ indicates whether the gripper is open or closed, $GraspStates = \{grasp, noGrasp\}$ indicates whether the robot is grasping the object or not, and $\Phi_{obj} \subseteq \mathbb{R}^6$ is the set of poses of the object in the robot's workspace. The action space is defined as $\actSpace = \actSpace_{\theta} \times \{openGripper, closeGripper\}$ where $\actSpace_{\theta} \subseteq \mathbb{R}^6$ is the set of fixed joint angle increments/decrements for each joint, and $openGripper, closeGripper$ are actions to open/close the gripper, resulting in $66$ actions. When executing a joint angle increment/decrement action $\hat{\theta}$, the joint angles evolve linearly according to $\theta_{t+1} = \theta_t + \Delta t \hat{\theta} + v_t$, 
where $\Delta t = 0.25$ and $v_t$ 
is a multivariate zero-mean Gaussian control error. We assume that the $openGripper$ and $closeGripper$ are deterministic.

Here the robot has access to two sensors: A joint-encoder that measures the joint angles of the robot and a grasp detector that indicates whether the robot grasps the object or not. For the joint-encoder, we assume that the encoder readings are disturbed by a small additive error drawn from a uniform distribution $[-0.05, 0.05]$. For the grasp detector we assume that we get a correct reading $90\%$ of the time. The robot starts from an initial belief where the gripper is open, the joint angles of the robot are $(0.8, -0.2, 0.8, -0.03, 0.0, 0.7)rad$ and the object is placed on the table such that the $x$ and $y$ positions of the object are uniformly distributed according to $[0.86m \pm 0.01m , 0.2 \pm 0.01m]$. When the robot collides with the environment or the object, it enters a terminal state and receives a penalty of -250. In case the robot closes the gripper but doesn't grasp the object, it receives a penalty of -100. Additionally, when the gripper is closed and a grasp is not established, the robot receives a penalty of -700 if it doesn't execute the $openGripper$ action. Each motion also incurs a small penalty of -3. When the robot successfully grasps the object, it receives a reward of 1,750 and enters a terminal state. The discount factor is 0.99 and we allow a maximum of 200 planning steps.

Similarly to the CarNavigation problem, the difficulty for this problem is the large number of steps that are required for the robot to complete its task (around 100). Additionally, the robot must act strategically when approaching the object to ensure a successful grasp.
\vspace{-0.25cm}
\subsection{Experimental setup}
All four test scenarios and the solvers are implemented in C++ within the OPPT framework\cite{hoerger2018software}, ensuring that all solvers use the same problem implementation. For ABT we used the implementation provided by the authors\cite{klimenko2014tapir}. For POMCP we used the implementation provided by \url{https://github.com/AdaCompNUS/despot}. Note that all three solvers rely on heuristic estimates of the action values in their rollout strategy. For a fair comparison, we use the same heuristic function for all three solvers, where we use methods from motion-planning, assuming the problem is deterministic.

All simulations were run single-threaded on an Intel Xeon Silver 4110 CPU with 2.1GHz and 128GB of memory.
For the \scene{4DOF-Factory} and \scene{KukaOffice} problem, we use the ODE physics engine\cite{Smi07:ODE} to simulate the transition dynamics. The levels $l$ used by \alg in these scenarios are associated with the ``discretization" (i.e., $\delta t$) used by the numerical integration of ODE. In particular, $\delta t = C_1 \cdot 2^{-C_2l}$. For the scenarios \scene{CarNavigation} and \scene{MovoGrasp}, since the dynamics of these problems are simple, \alg associates the levels $l$ to the time-step, i.e., $\Delta t = C_1 \cdot 2^{-C_2l}$. The exact parameters (i.e., $C_1$, $C_2$, and the number of levels $L$) were determined via systematic preliminary trials. As a result of these trials, we set the parameters used by \alg for \scene{4DOF-Factory} and \scene{KukaOffice} to be $C_1$=$0.0128$, $C_2$=$1$, $L$=$7$, for \scene{CarNavigation} to be $C_1$=$0.4$, $C_2$=$1$, $L$=$3$, and for \scene{MovoGrasping} to be $C_1$=$1$, $C_2$=$0.5$, $L$=$4$.

The purpose of our experiments are three folds. First is to test whether our particular choice for the multiple levels of approximation of the transition functions results in a reduction of the variance of the $Q$-value difference terms in \eref{eq:correction_term}. This ensures that, as we increase the level, fewer and fewer episode samples are required to accurately estimate the difference terms. To do this, we ran \alg on each problem scenario for 10 runs with a planning time of 20s per step. Then, at each step, after planning time is over, we use the computed policy $\pi$ and sample 50,000 additional episodes from the current history \hist on each level $l$ to compute the variance $\mathbb{V}[Q_l(\hist, \act)]$ and 50,000 correlated episodes on each level $l$ to compute $\mathbb{V}[Q_{l}(\hist, \act) - Q_{l-1}(\hist, \act)]$, where \act is the action performed from \hist according to $\pi(\hist)$. Taking the average of these variances over all steps and all simulation runs then gives us an indication how the variance of the $Q$-value difference terms in \eref{eq:correction_term} behaves as we increase the level of approximation of the transition function.

Second is to compare \alg with two state-of-the-art POMDP solvers ABT \cite{kurniawati2016online} and POMCP\cite{silver2010monte}. For this purpose, we used a fixed planning-time per step for each solver, where we used 1s for the \scene{4DOF-Factory}, \scene{CarNavigation} and \scene{MovoGrasp} problem, and 5s for the \scene{KukaOffice} problem. For each problem scenario we tested ABT and POMCP using different levels of approximations of \transF for planning, to see whether using a single approximation of \transF helps to speed-up computing a good policy,
compared to MLPP that uses all levels of approximations of $T$ for planning.

DESPOT\cite{somani2013despot} is not used as a comparator because for the type of problems we try to address, DESPOT's strategy of expanding each belief with every action branch (via forward simulation) is uncompetitive. For example, for \scene{4DOF-Factory}, expanding a single belief takes, on average, $\sim$14.4s using K=50 scenarios (50 is a tenth of what it commonly used\cite{somani2013despot}), which is already much more than the time for a single planning step in our experiments (1s). Similarly, for the long planning-horizon problem \scene{MovoGrasp}, DESPOT must expand all 66 actions using K scenarios from every belief it encounters, which quickly becomes infeasible for a planning horizon of more than 5 steps.

Last, we investigated if and how fast \alg converges to a near-optimal policy compared to ABT and POMCP, when the latter two solvers use the original transition function for planning. To do this, we used multiple increasing planning times per step for the \scene{4DOF-Factory} problem, starting from 1s to 20s per step.  
The results of all three experiments are discussed in the next section.
\vspace{-0.35cm}
\subsection{Results}
\vspace{-0.25cm}
\subsubsection{Variances of $Q_{l} - Q_{l-1}$}\label{sssec:variance}
\begin{figure}
\centering
\vspace{-0.5cm}
\begin{tabular}{c@{\hskip0pt}c@{\hskip0pt}c@{\hskip0pt}c}
\includegraphics[width=0.25\textwidth]{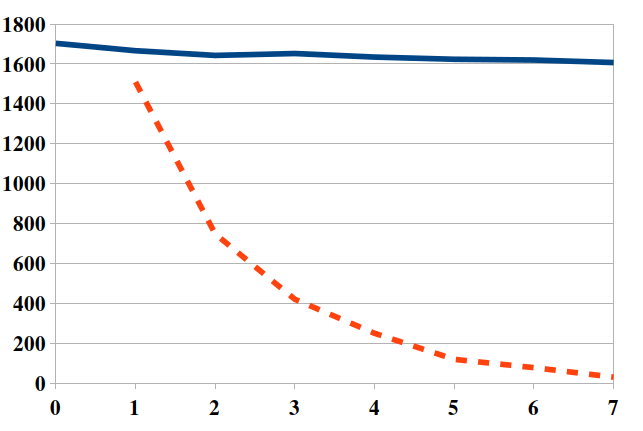} &
\includegraphics[width=0.25\textwidth]{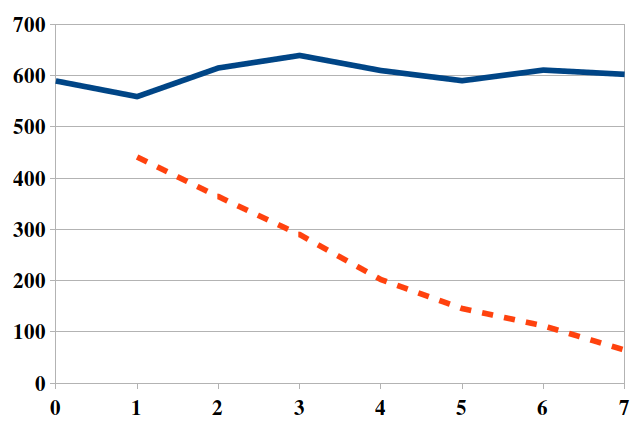} &
\includegraphics[width=0.25\textwidth]{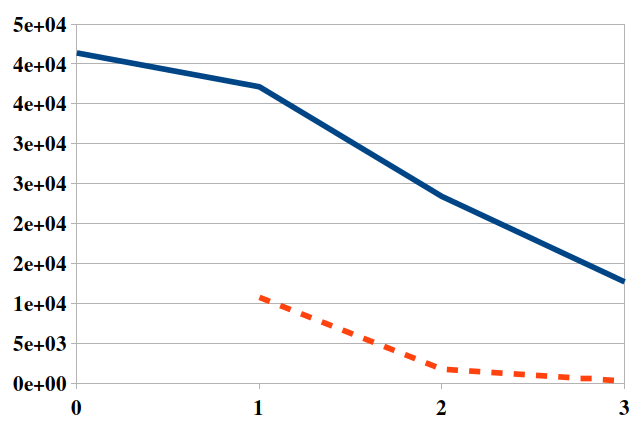} &
\includegraphics[width=0.25\textwidth]{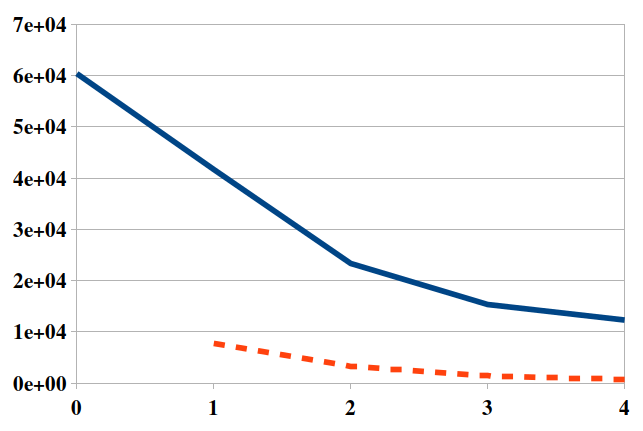} \\
(a) 4DOF-Factory & (b) KukaOffice & (c) CarNavigation & (d) MovoGrasp
\end{tabular}
\caption{Average variance of $Q_l$ (solid blue line) and $Q_l - Q_{l-1}$ (dashed red line) for the problem scenarios (a) 4DOF-Factory, (b) KukaOffice, (c) CarNavigation and (d) MovoGrasp. The $x$-axis represents the level $l$ and the $y$-axis represents the variance.}
\label{f:variances}
\vspace{-1.25cm}
\end{figure}

Figure \ref{f:variances} shows the average variances of $Q_l$ and $Q_{l} - Q_{l-1}$ for all four problem scenarios. It is clear that in all scenarios the variance of the $Q$-value differences decreases significantly as we increase the level $l$, indicating that we indeed require fewer and fewer episode samples for increasing $l$. Note that the rate of decrease depends on the particular choice of the sequence of approximate transition functions. Multiple sequences can be possible for a particular problem, but preference should be given to the sequence for which the variance of the $Q$-value difference decreases fastest. 

\subsubsection{Average total discounted rewards}\label{sssec:avgRewards}
\begin{figure}
\vspace{-0.5cm}
\centering
\small
\begin{tabular}{c@{\hskip0pt}c@{\hskip0pt}c@{\hskip0pt}c}
\includegraphics[width=0.25\textwidth]{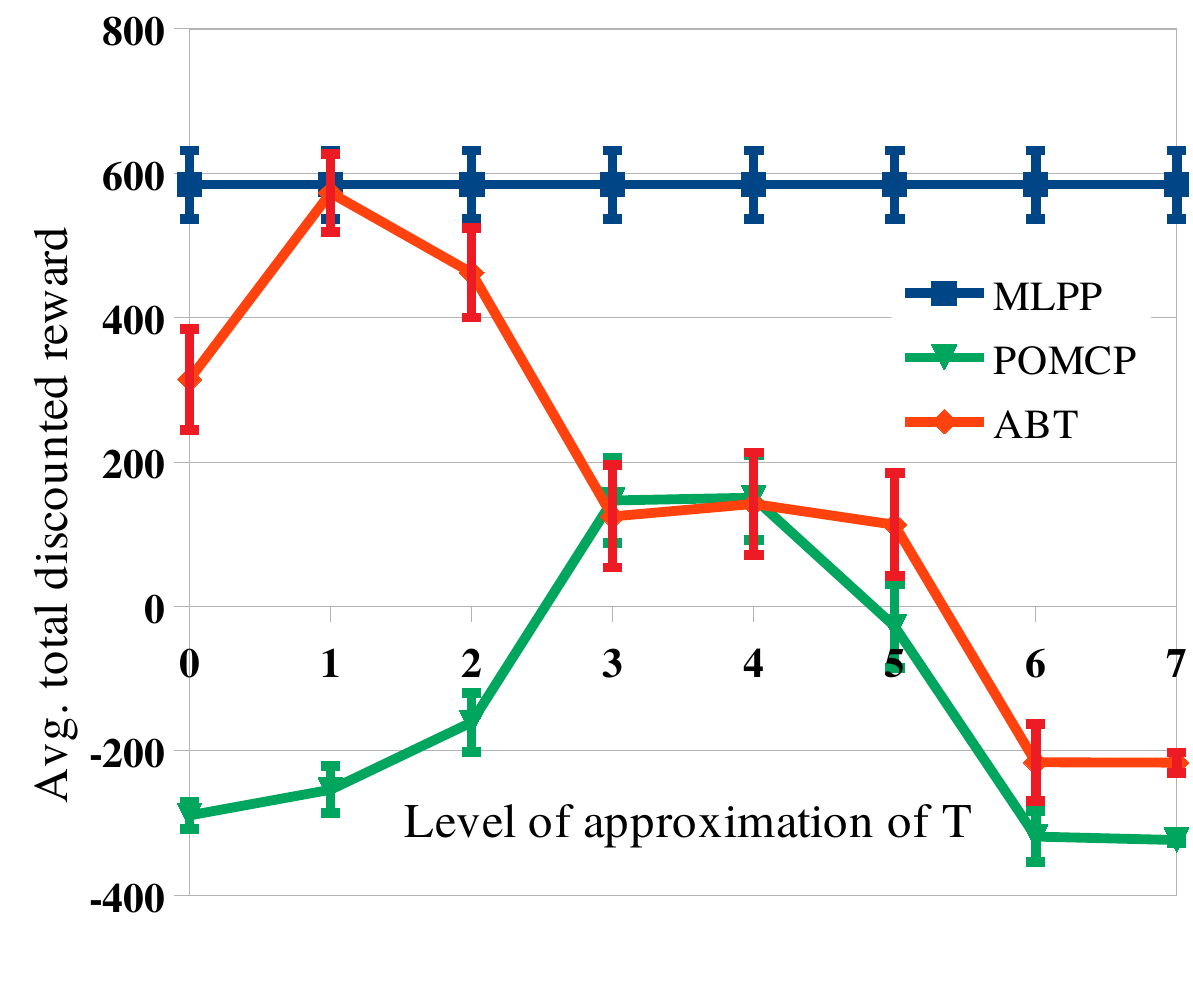} &
\includegraphics[width=0.25\textwidth]{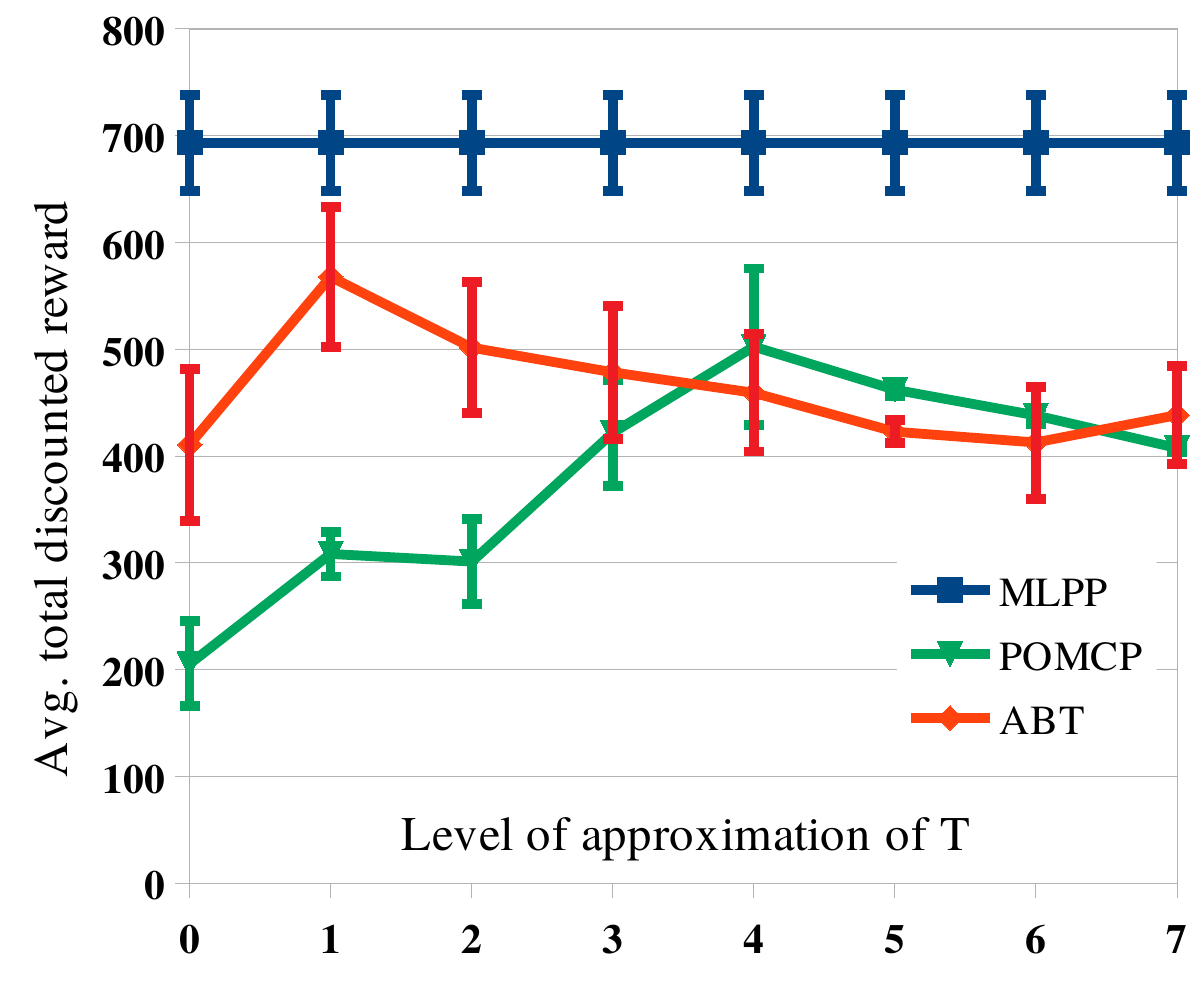} &
\includegraphics[width=0.25\textwidth]{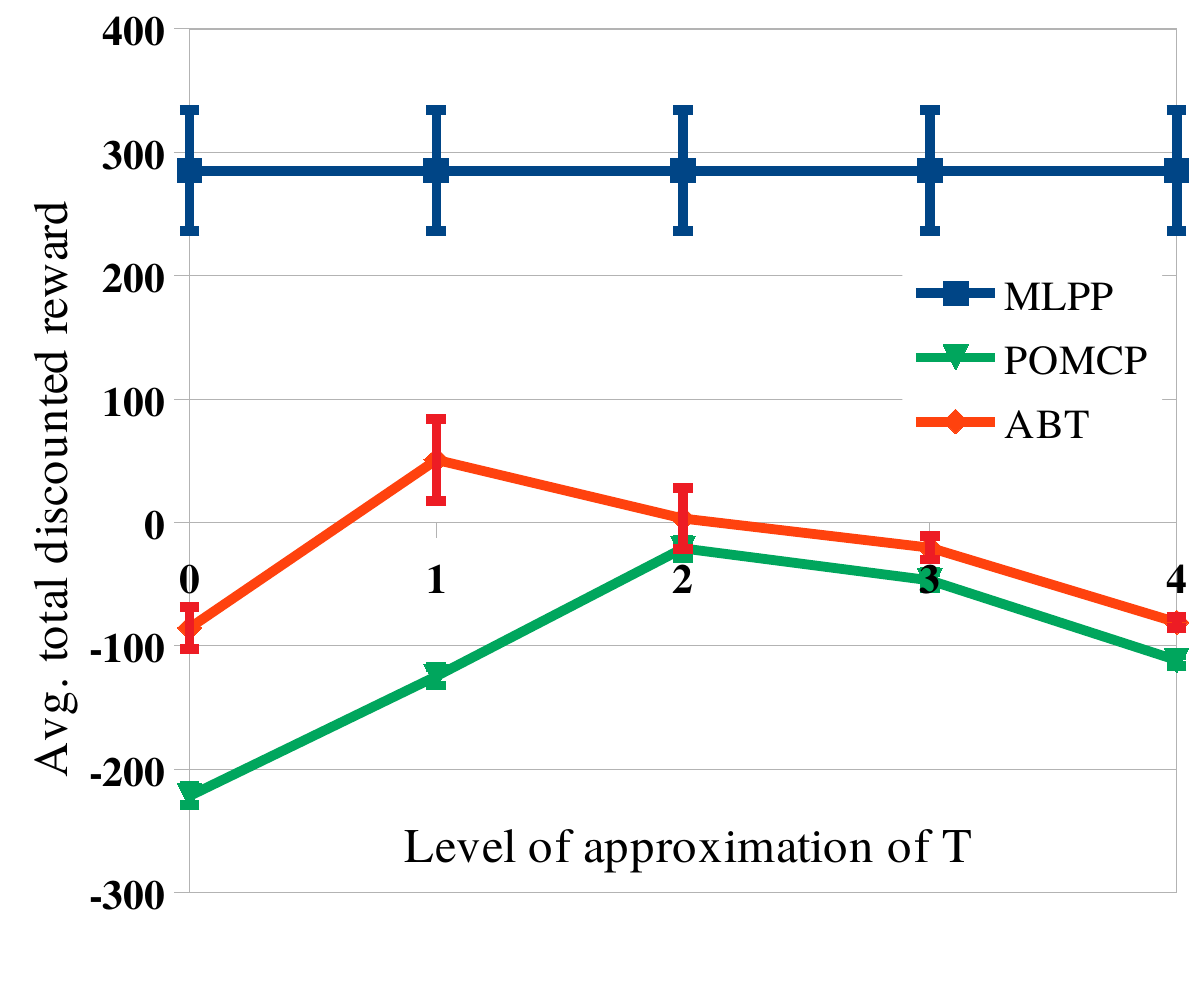} &
\includegraphics[width=0.25\textwidth]{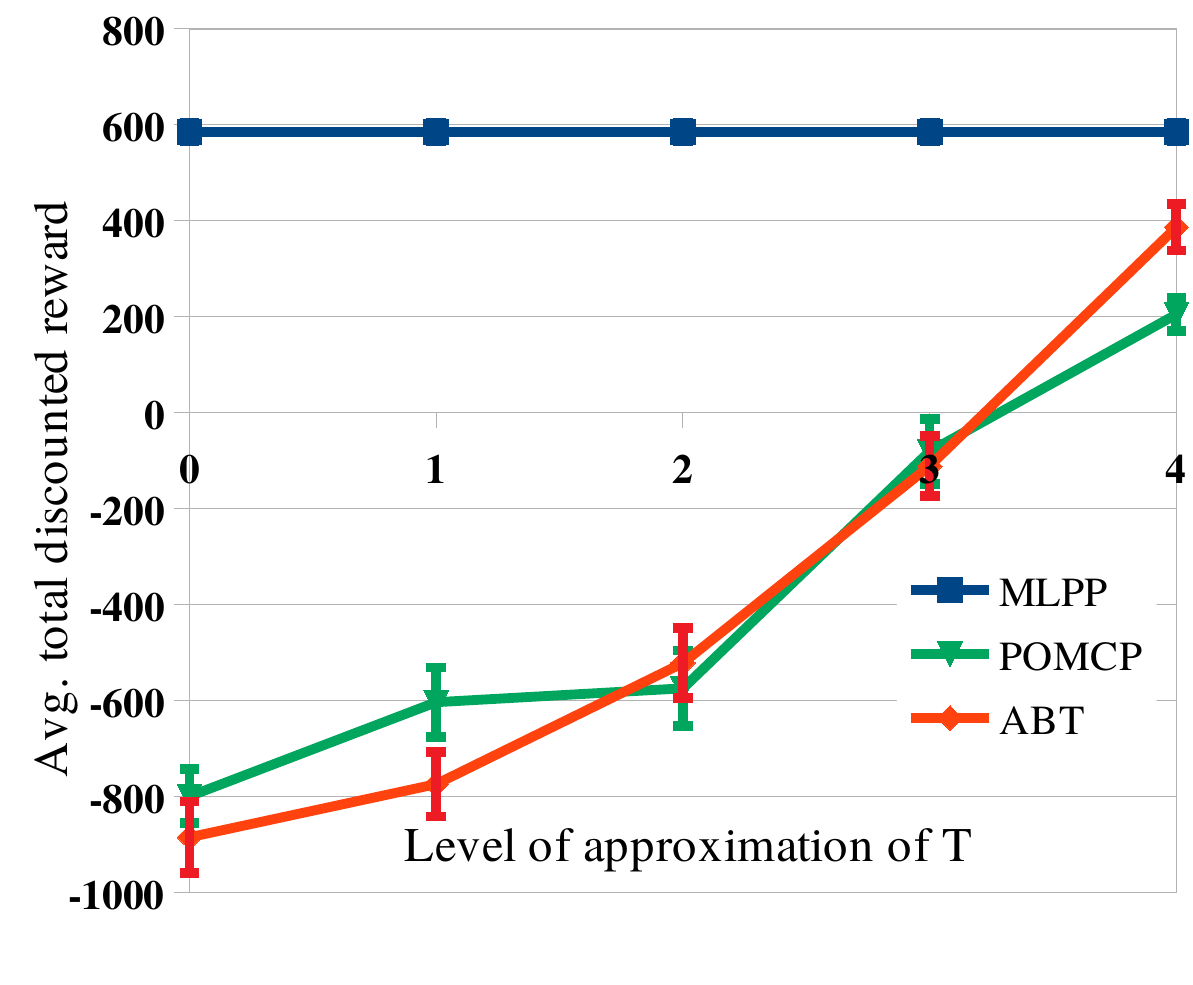} \\
(a) & (b) & (c) & (d)
\end{tabular}
\caption{Average total discounted reward of MLPP, ABT and POMCP on the \scene{4DOF-Factory} (a), \scene{KukaOffice} (b), \scene{CarNavigation} (c) and \scene{MovoGrasp} (d) scenarios. The $x$-axis represents the level of approximation of the transition function used for planning. Note that MLPP uses all levels for planning (hence the horizontal lines), whereas ABT and POMCP use only a single level as indicated by the $x$-axis. For each scenario, the largest level of approximation is equal to the original transition function. 
Vertical bars are the 95\% confidence intervals.}
\label{f:plots}
\vspace{-0.5cm}
\end{figure}

\fref{f:plots}(a)-(d) shows the average total discounted rewards achieved by ABT, POMCP and \alg in all four test scenarios. The results indicate that, for ABT and POMCP, using a single coarse approximation of \transF for planning can help compute a better policy, compared to using the original transition function. However, different regions of the belief space are likely to require different level of approximations. For instance in \scene{4DOF-Factory}, when the states in the support of the belief place the robot in the relatively open area, coarse levels of approximation suffice but, when they are in the cluttered area, higher accuracy is required. Unlike ABT and POMCP, MLPP covers multiple levels of approximations and is able to quickly reduce errors in the estimates of the action values caused by coarse approximations. 
The lack of coverage causes difficulties for ABT and POMCP 
in \scene{MovoGrasping} as well, where a high accuracy is neccessary for grasping. 

\vspace{-0.5cm}
\begin{wrapfigure}{L}{0.5\textwidth}
\vspace{-1cm}
\includegraphics[width=0.5\textwidth]{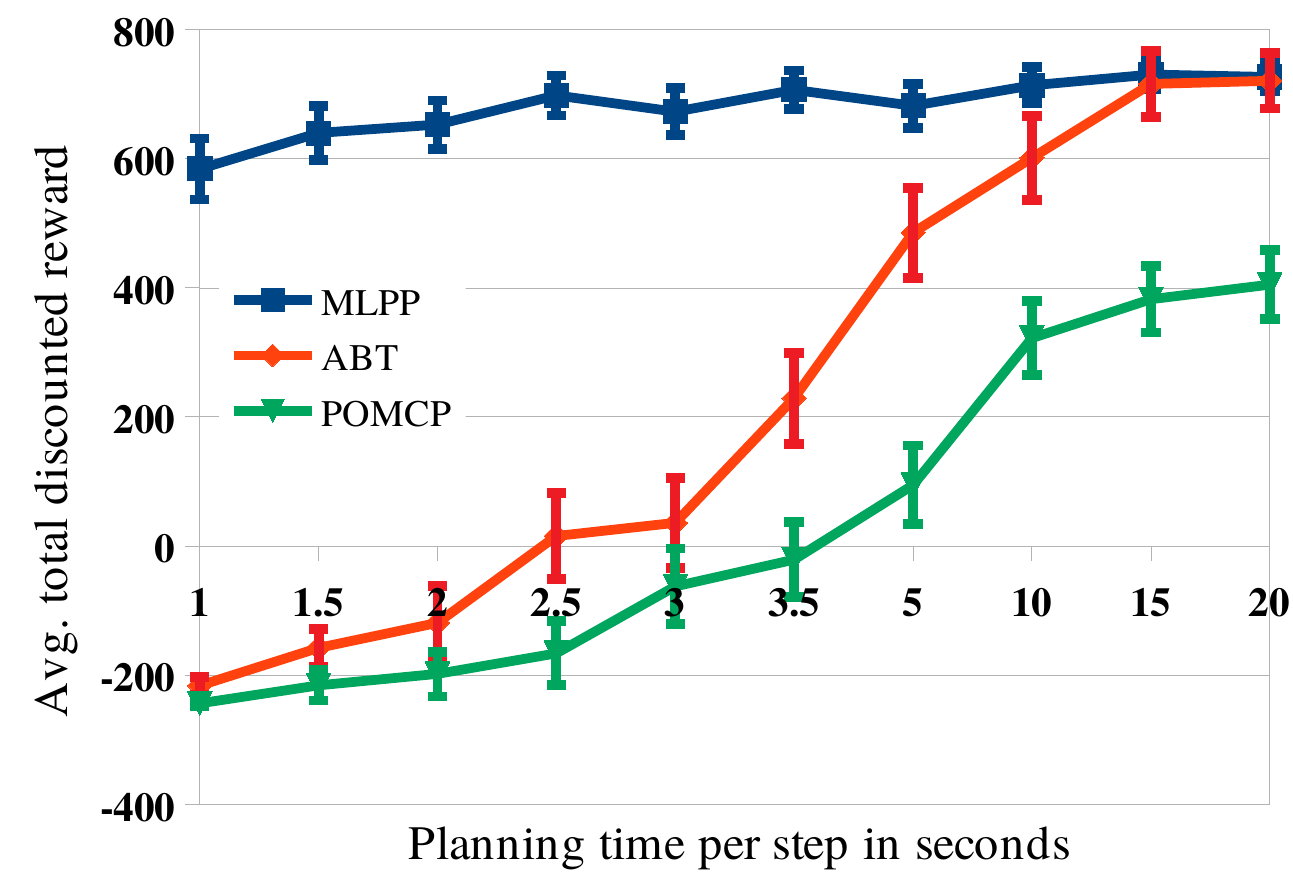}
\vspace{-0.5cm}
\caption{Average total discounted rewards for ABT, POMCP and \alg for \scene{4DOF-Factory} using increasing planning times per step. The average is taken over 500 simulation runs for each planning time and algorithm. Vertical bars are the 95\% confidence intervals.}
\label{f:4DOFScenario}
\vspace{-1.0cm}
\end{wrapfigure}

\subsubsection{Increasing planning times}\label{sssec:convergence}
Figure \ref{f:4DOFScenario} shows the average total discounted rewards achieved by each solver for the \scene{4DOF-Factory} scenario as the planning time per step increases. The results indicate \alg converges to a good policy much faster than ABT and POMCP: ABT requires 15s/step to generate a policy whose quality is similar to the policy generated by \alg in 2.5s/step, while POMCP is unable to reach similar level of quality, even with a planning time of 20s/step (in our experiments it takes roughly 5 minutes of planning time/step for POMCP to converge to a near-optimal policy).

\section{Conclusion}\label{sec:conclusion}
Despite the rapid progress in on-line POMDP planning, computing robust policies for systems with complex dynamics and long planning-horizons remains challenging. Today's fastest on-line solvers rely on a large number of forward simulations and standard Monte-Carlo methods to estimate the expected outcome of action sequences. While this strategy works well for small to medium-sized problems, their performance quickly deteriorates for problems with transition dynamics that are expensive to evaluate and problems with long planning-horizons.

To alleviate these shortcomings, we propose \alg, an on-line POMDP solver that extends Multilevel Monte-Carlo to POMDP planning. \alg samples histories using multiple levels of approximation of the transition function and computes an approximation of the action-values using a Multilevel Monte-Carlo estimator. This enables \alg to significantly speed-up the planning process while retaining correctness of the action-value estimates. We have successfully tested \alg on four robotic tasks that involve expensive transition dynamics and long planning-horizons. In all four tasks, \alg outperforms ABT and POMCP, two of the fastest on-line solvers, which shows the effectiveness of the proposed method.

%
%
\bibliographystyle{spmpsci}

\end{document}